\documentclass{article}
\usepackage[main,final]{neurips_2026}
\usepackage[utf8]{inputenc}
\usepackage[T1]{fontenc}
\usepackage{hyperref}
\usepackage{url}
\usepackage{booktabs}
\usepackage{amsfonts,amsmath,amssymb}
\usepackage{nicefrac,microtype,xcolor,graphicx}
\usepackage{array,multirow,enumitem,float,longtable}
\usepackage{tabularx}
\usepackage{placeins}
\usepackage{needspace}
\hypersetup{hidelinks}
\bibpunct{[}{]}{,}{a}{,}{,}

\newcommand{\papertablestyle}{%
  \footnotesize
  \renewcommand{\arraystretch}{1.15}%
  \setlength{\tabcolsep}{3pt}%
  \setlength{\aboverulesep}{0.6ex}%
  \setlength{\belowrulesep}{0.6ex}%
}
\AddToHook{env/table/begin}{%
  \setlength{\abovecaptionskip}{6pt}%
  \setlength{\belowcaptionskip}{6pt}%
}
\title{Is Text All You Need? Text as a Universal Information Bottleneck for Speech LLMs}
\author{%
  Ming-Hao Hsu$^{1,\dagger}$ \quad
  Yuxuan Hu$^{2}$ \quad
  Shujie Liu$^{3,*}$ \quad
  Jinyu Li$^{2}$ \quad
  Yan Lu$^{3}$ \quad
  Zhizheng Wu$^{1,*}$
}
\begin{document}
\maketitle
\begingroup
\renewcommand{\thefootnote}{}
\footnotetext{$^{1}$School of Data Science, The Chinese University of Hong Kong, Shenzhen, China. $^{2}$Microsoft Research, Redmond, WA, USA. $^{3}$Microsoft Research Asia, Hong Kong, China. $^*$Correspondence to: Shujie Liu $<$\texttt{shujliu@microsoft.com}$>$ and Zhizheng Wu $<$\texttt{wuzhizheng@cuhk.edu.cn}$>$. $^\dagger$Work done during an internship at Microsoft.}
\endgroup
\begin{abstract}
Large language models (LLMs) provide a powerful reasoning backbone for speech understanding, but integrating continuous acoustic signals into a pretrained LLM remains challenging.
A central design question is how to tie speech representations to the LLM's text input space while preserving the continuous variation needed for different speech tasks.
In this work, we propose Convex Gate (C-Gate), a speech--LLM bridge that represents each frame as a convex combination of frozen token embeddings.
This architectural convex-hull constraint fixes the output codebook while allowing continuous mixtures learned from task supervision.
With a Whisper-large-v3 encoder, a partially adapted Qwen2.5-7B-Instruct decoder, and joint training on speech recognition, emotion recognition, and spoken reasoning, C-Gate performs accurate transcription and emotion recognition, our representative semantic and acoustic tasks, through one shared decoder.
On text-only benchmarks, the adapted decoder loses instruction-following and MMLU accuracy and matches or exceeds the original model on GSM8K and HellaSwag.
We further design analyses and interventions on the trained model to examine where the bridge sequence carries task information and how the decoder reads it.
These findings show how fixed text embeddings can support both semantic and acoustic readouts through continuous speech-conditioned mixtures.
\end{abstract}

\section{Introduction}
\label{sec:intro}

Reusing pretrained large language models as the reasoning backbone for speech understanding is increasingly necessary for scalable multimodal systems.
Instruction-tuned LLMs such as Qwen2.5-7B-Instruct \citep{yang2024} carry world knowledge, calibrated answer behavior, and tool-use priors whose retraining from speech-aligned data alone would be prohibitively expensive.
A standard design pattern attaches a speech encoder to a pretrained LLM via a learned bridge \citep{li2023,rubenstein2023,zhang2023,tang2024,chu2023,chu2024}, allowing a single decoder to serve transcription, paralinguistic, and spoken-reasoning workloads.
Under this paradigm, the central design question is what representation form should be used to feed speech information into the LLM context.
One answer is continuous prompting, which conditions pretrained LLMs through learned prefix representations \citep{li2021,lester2021}.
Another, increasingly popular, answer constructs speech representations from the LLM's own input embedding table \citep{ma2025,yang2025}.

Existing interfaces make different choices about vocabulary alignment and continuous expressivity.
Some methods align speech representations with discrete tokens using objectives such as CTC or text alignment \citep{ma2025,fan2025,peng2026,wang2026a}.
These objectives emphasize semantic content, while emotion recognition also requires acoustic cues such as prosody.
Other approaches learn continuous tokens or latent embeddings without restricting them to mixtures of vocabulary embeddings \citep{tang2024,chu2023,chu2024}.
This gives the bridge flexibility to choose its output representations.
Together, these designs motivate separating two choices: the embedding vectors available to the bridge and the supervision used to combine them.
We ask whether a fixed text embedding table can support both semantic and acoustic targets when speech selects continuous mixtures through task supervision.
We take speech recognition and emotion recognition as the representative semantic and acoustic tasks.

To study this question, we propose the Convex Gate (C-Gate), a speech--LLM bridge with an architectural convex-hull constraint.
We represent each time step as a convex combination of LLM input embeddings, fixing the output codebook while preserving continuous expressivity.
Concretely, C-Gate scores the full vocabulary using downsampled speech features, selects the top-16 entries, and renormalizes their weights to combine rows of the frozen embedding table $\mathbf{E}$.
The bridge uses the raw embedding values, with no value projection or post-codebook multilayer perceptron (MLP).
This construction separates a fixed output geometry from the choice of routing supervision.
Within this construction, task losses train only the selection and the mixture weights, allowing the support to carry information useful for transcription or emotion recognition.
The shared embedding coordinates also make the selected tokens, mixture weights, and decoder attention directly inspectable.

Our experiments show that C-Gate supports both transcription and emotion recognition with a shared decoder.
Beyond task performance, we find that selected token identities carry emotion information even when their mixture weights are nearly uniform.
Interventions on trained models further show that decoding depends on temporal order and the learned key--value correspondence.
These findings distinguish the information recoverable from the selected supports from how the decoder uses the full bridge sequence.

We summarize our contributions as follows:

\begin{itemize}[leftmargin=*,itemsep=3pt]
\item \textbf{Geometry-Constrained Speech--LLM Interface:} We introduce C-Gate, which represents each speech frame as a sparse convex mixture of 16 rows of the frozen LLM embedding table, selected by scoring the full vocabulary.
The bridge applies no projection after the embedding table, so its outputs stay inside the convex hull of the text embeddings, and within the bridge the task losses train only which rows are selected and how they are weighted.
\item \textbf{Semantic and Acoustic Tasks from One Frozen Codebook:} We show that convex mixtures of a frozen text embedding table support transcription as a semantic task and emotion recognition as an acoustic task with a shared, partially adapted decoder.
In our experiments, a single checkpoint trained jointly on speech recognition, emotion recognition, and spoken reasoning reaches 3.98\% and 8.42\% word error rate (WER) on LibriSpeech test-clean and test-other and 84.3\% accuracy on emotion test recordings from eight corpora.
\item \textbf{Representation and Readout Analysis:} We show that the mixture weights are nearly uniform and that the identities of the selected tokens alone support a linear emotion readout at least as accurate as that of the pooled encoder states.
Interventions on the trained model show that decoding depends on frame order and on the learned key--value correspondence, while setting the weights to exactly uniform has a much smaller effect.
\end{itemize}

\section{Related Work}
\label{sec:related}

\paragraph{Speech-to-LLM Adapters.}
Speech--LLM systems require an adapter that converts acoustic encoder features into a sequence of embeddings the language-model decoder can consume.
Adapter designs differ in how they aggregate acoustic features over time and construct the representations passed to the decoder \citep{tang2024,ma2024a,lee2026}.
One line of designs uses query-based aggregation, which has a precedent in vision--language modeling \citep{alayrac2022}, where BLIP-2 introduced the Q-Former to connect a frozen image encoder and a frozen language model \citep{li2023}.
SALMONN extends this design to speech and audio through a window-level Q-Former that produces a variable-length sequence of audio embeddings \citep{li2023,tang2024}.
Qwen-Audio and Qwen2-Audio, in turn, connect audio encoders to LLMs through audio--text pretraining across multiple tasks \citep{chu2023,chu2024}.
However, these adapters leave the bridge a free output space, so where its outputs settle relative to the text embeddings is an outcome of training \citep{hsu2026a} rather than a property of the design.
We instead fix that geometry by construction and ask what a bridge confined to the text embedding hull can still carry.

\paragraph{Vocabulary Mixtures and Alignment Objectives.}
Beyond the adapter architecture, a speech--LLM interface is shaped by both its output representation and the supervision used to learn the mapping \citep{wang2023,deng2025,ma2025}.
The output representation can also be discrete, and codec-based tokenizers such as SpeechTokenizer show that discrete speech units can retain speaker and prosodic information alongside content \citep{zhang2024,mousavi2024,hsu2025}.
Alignment can guide temporal aggregation, as in SSR's segmentation of speech features and AlignFormer's use of CTC alignments to define dynamic Q-Former windows \citep{tan2025,fan2025}.
CTC posteriors can also provide a common interface for speech and text \citep{ma2025}, which TASU exploits by simulating these posteriors from text-only data during alignment training \citep{peng2026}.
Behavioral supervision offers another route \citep{cuervo2025}: BLSP learns from text continuations, while TARS uses reinforcement learning to align speech- and text-conditioned hidden representations and generated outputs \citep{wang2023,wang2026a}.
Yet in each case the tie between speech and text enters through the training signal, and a transcript or a text-conditioned behavior defines what the speech representation should match.
By contrast, C-Gate places the tie in the architecture and trains the bridge with the task losses alone, so the targets decide what the mixture keeps.

A direct connection to the LLM vocabulary can also be built into the output representation by constructing mixtures of token embeddings \citep{ma2025,yang2025}.
LegoSLM uses CTC posteriors as mixture weights over LLM embeddings, with an additional blank representation, to reconstruct speech embeddings \citep{ma2025}.
For automatic speech recognition (ASR), \citet{yang2025} softly discretize audio using a codebook initialized from the LLM embedding table, studying sparse mixtures with both updated and fixed codebooks.
These methods establish vocabulary mixtures as a speech--LLM interface for transcription, where the target is itself a token sequence and a CTC posterior or an ASR-trained discretization supplies the weights \citep{ma2025,yang2025}.
They leave open, however, whether the original frozen table can also carry an acoustic target that no transcript token names, with weights learned from the task losses alone.
C-Gate takes up this question: each bridge output is a normalized mixture over selected rows of the original, frozen embedding table, and we learn the mixture weights from task losses without CTC supervision.

\paragraph{Evaluation beyond Transcription.}
Transcription accuracy alone does not reveal how well a speech--LLM interface supports paralinguistic understanding and spoken reasoning \citep{ma2024b,wang2024a,wang2026b,wang2026c,hsu2026b}.
VoiceBench evaluates responses to spoken instructions \citep{chen2026}, and SpeechMMLU probes knowledge-based question answering from speech \citep{tan2025}.
MMAU extends evaluation to understanding and reasoning across speech, environmental sounds, and music \citep{sakshi2025}.
Interpreting these scores also requires distinguishing correct answers from answers that depend on the audio input \citep{rouditchenko2025,zang2025,he2026}.
Work on audio contribution examines this distinction directly \citep{he2026}, and studies of multiple-choice evaluation show that scores can vary with the ordering and wording of answer choices \citep{lpez2025}.
These controls ask whether a benchmark score depends on the audio.
Beyond adopting them for the reasoning benchmarks, we extend the same logic to the bridge and intervene on the waveform, the key--value mapping, and the frame order of a trained checkpoint to locate what the decoder reads for transcription and emotion recognition.

\section{Convex Gate}
\label{sec:method}

\begin{figure}[t]
\centering
\includegraphics[width=0.95\linewidth]{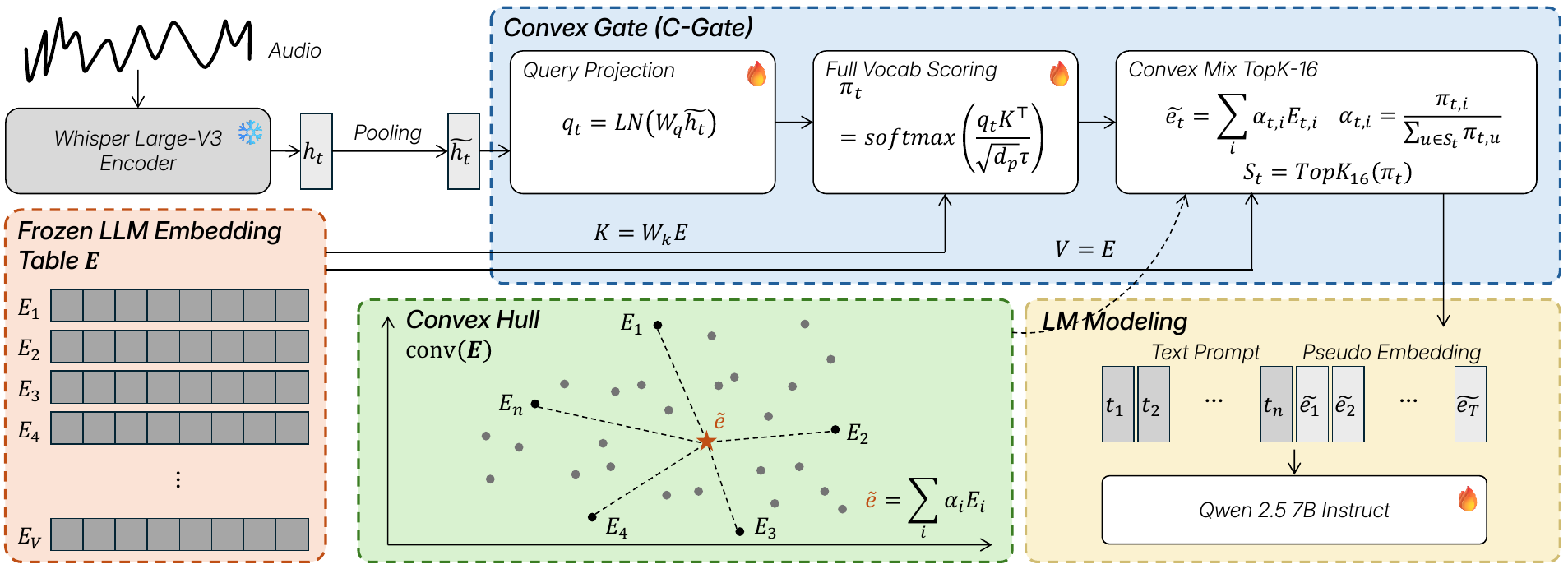}
\caption{\textbf{C-Gate Architecture.}
A sparse convex mixture of frozen token embeddings connects the speech encoder to the partially adapted decoder.
Snowflakes mark frozen modules, and flames mark components with trainable parameters.
Decoder adaptation is restricted to the attention projections specified in Section~\ref{sec:model}.}
\label{fig:1}
\label{fig:arch}
\end{figure}

C-Gate converts acoustic features into weighted averages of the LLM's input embeddings (Fig.~\ref{fig:1}).
The acoustic scorer determines which entries to combine, while the frozen embedding table determines the values passed to the decoder.

\paragraph{Geometric Constraint.}
Let $\mathbf{E}\in\mathbb{R}^{V\times d_{\mathrm{llm}}}$ be the frozen embedding table, whose row $\mathbf{E}_v$ is the embedding of token $v$.
The permitted outputs belong to
\begin{equation}
\operatorname{conv}(\mathbf{E})=\left\{\sum_{v=1}^{V}a_v\mathbf{E}_v:a_v\geq0,\ \sum_{v=1}^{V}a_v=1\right\}.
\label{eq:1}
\end{equation}
This construction fixes the output codebook while allowing continuous mixtures between its rows.
The resulting sequence can carry speech-dependent variation through both the selected embeddings and their mixture weights.

\paragraph{Acoustic Queries and Vocabulary Keys.}
The frozen encoder produces $\mathbf{h}_{1:T}\in\mathbb{R}^{T\times d_{\mathrm{enc}}}$.
Mean pooling with stride $r$ gives $T'=\lceil T/r\rceil$ pooled states $\tilde{\mathbf{h}}_{1:T'}$.
Using row vectors throughout, we form queries and keys as
\begin{equation}
\mathbf{q}_t=\operatorname{LN}(f_q(\tilde{\mathbf{h}}_t))\in\mathbb{R}^{1\times d_p},\qquad \mathbf{K}=f_k(\mathbf{E})\in\mathbb{R}^{V\times d_p},
\label{eq:2}
\end{equation}
where $f_q$ and $f_k$ are learned projection maps with weight matrices $\mathbf{W}_q\in\mathbb{R}^{d_p\times d_{\mathrm{enc}}}$ and $\mathbf{W}_k\in\mathbb{R}^{d_p\times d_{\mathrm{llm}}}$, applied to row vectors by right multiplication with their transposes.
Layer normalization acts on the query, and the key projection enters only the scores, so it changes which entries are selected while leaving the values unchanged.
With a positive learned temperature $\tau$, we score the vocabulary and select the largest $k$ entries:
\begin{equation}
\mathbf{z}_t=\frac{\mathbf{q}_t\mathbf{K}^\top}{\sqrt{d_p}\,\tau},\qquad \boldsymbol{\pi}_t=\operatorname{softmax}(\mathbf{z}_t),\qquad S_t=\operatorname{TopK}_k(\boldsymbol{\pi}_t),
\label{eq:3}
\end{equation}
where $S_t\subset\{1,\dots,V\}$ indexes the $k$ largest entries of $\boldsymbol{\pi}_t$.

\paragraph{Raw Embedding Mixtures.}
The selected probabilities are renormalized before constructing each bridge output:

\begin{equation}
\alpha_{t,v}=\begin{cases}
\displaystyle\frac{\pi_{t,v}}{\sum_{u\in S_t}\pi_{t,u}},&v\in S_t,\\[4pt]
0,&v\notin S_t,
\end{cases}
\qquad\tilde{\mathbf{e}}_t=\sum_{v\in S_t}\alpha_{t,v}\mathbf{E}_v\in\operatorname{conv}(\mathbf{E}).
\label{eq:4}
\end{equation}
Nonnegative coefficients summing to one establish the inclusion in Eq.~\ref{eq:1} directly.
The mixture is passed to the decoder as is, with no learned value projection or post-mixture MLP.
For training, support selection is discrete, while the selected weights are differentiable wherever the support remains unchanged.
The selection receives no supervision of its own, such as a transcript-token target, and is shaped only by the task losses below.

\paragraph{Decoder Interface.}
The decoder conditions on the task instruction $\mathbf{x}$ and the bridge sequence $\tilde{\mathbf{e}}_{1:T'}$ and generates the target tokens $y_{1:N}$ autoregressively:

\begin{equation}
p_{\mathcal M}(y_{1:N}\mid \mathbf{x},\tilde{\mathbf{e}}_{1:T'})=\prod_{i=1}^{N}p_{\mathcal M}\bigl(y_i\mid\mathcal{C}(\mathbf{x},\tilde{\mathbf{e}}_{1:T'}),y_{<i}\bigr),
\label{eq:5}
\end{equation}
where $\mathcal{C}$ denotes the prompt and audio context rendered with the decoder's chat template.
The text tokens of the prompt use the same frozen embedding table.
At inference, the same model generates transcripts, emotion completions, or reasoning answers.
It uses no separate emotion classifier or task-specific decoder.

\paragraph{Training Objective.}
We train the acoustic scorer and selected decoder attention parameters by minimizing the expected next-token cross-entropy for each task:

\begin{equation}
\mathcal{L}_{\mathrm{CE}}=\mathbb{E}_{\mathcal D}\left[-\log p_{\mathcal M}\bigl(y_i\mid\mathcal{C}(\mathbf{x},\tilde{\mathbf{e}}_{1:T'}),y_{<i}\bigr)\right].
\label{eq:6}
\end{equation}
Here $\mathcal D$ denotes the distribution of target-token prediction examples for the task, and $y_{<i}$ is the ground-truth target prefix.
When several tasks are trained jointly, we sum their losses with dynamic weights that follow each task's loss relative to its initial value (Appendix~\ref{app:training}).
Throughout training, the speech encoder and the embedding values remain frozen.
Section~\ref{sec:setup} specifies the model configuration and adaptation scope.

\section{Experimental Setup}
\label{sec:setup}

This section describes the model configuration, the training data and optimization, and the evaluation benchmarks.
More details can be found in Appendices~\ref{app:training} and~\ref{app:eval}.

\subsection{Model Configuration}
\label{sec:model}

We initialize the acoustic encoder with pretrained Whisper-large-v3 \citep{radford2023} and the autoregressive decoder with Qwen2.5-7B-Instruct \citep{yang2024}.
Following the notation in Section~\ref{sec:method}, the encoder produces acoustic features $\mathbf{h}_t$ with encoder dimension $d_{\mathrm{enc}}=1{,}280$.
The decoder's input embedding table $\mathbf{E}$ has embedding dimension $d_{\mathrm{llm}}=3{,}584$ and $V=152{,}064$ rows.
The C-Gate scorer maps acoustic features and vocabulary embeddings into queries $\mathbf{q}_t$ and keys $\mathbf{K}$ with a shared scoring dimension $d_p=512$.
We set the support size $k$ to 16 and the mean-pooling stride $r$ to 4.

During adaptation, we train the C-Gate scorer and the query, key, value, and output attention projections in decoder layers 0--23.
These are 2.5M scorer parameters and 704.8M attention parameters, which together make up 8.6\% of the 8.26B parameters of the full model.
All remaining parameters are frozen, including the Whisper-large-v3 encoder, the input embeddings, MLPs, normalization layers, language-model head, and decoder layers 24--27.
The embedding values mixed by C-Gate therefore remain fixed throughout training.

\subsection{Training Data and Optimization}
\label{sec:training}

\paragraph{ASR.}
We train on the 960 hours of LibriSpeech \citep{panayotov2015}.

\paragraph{Emotion Recognition.}
We pool the English portion of CAMEO \citep{christop2025}, which includes CREMA-D, EMNS, eNTERFACE, JL-Corpus, and RAVDESS, with ESD English \citep{zhou2021}, IEMOCAP \citep{busso2008}, SAVEE \citep{jackson2014}, TESS \citep{dupuis2011}, and MELD audio \citep{poria2019}, and map their labels to eight classes.
We split each corpus by recording into a training and a test portion.
We train on 53,246 recordings, and the emotion test split has 2,063 recordings from CREMA-D, EMNS, eNTERFACE, JL-Corpus, ESD, IEMOCAP, SAVEE, and TESS.

\paragraph{Reasoning.}
We use question-answer corpora with synthesized spoken questions, including ARC \citep{clark2018}, GSM8K \citep{cobbe2021}, MATH \citep{hendrycks2021b}, and MMLU-Pro \citep{wang2024b}.
Appendix~\ref{app:training} lists all sources and their sizes.

\paragraph{Training Configurations.}
C-Gate denotes the three-task model trained jointly on ASR, emotion recognition, and reasoning.
The other training-task combinations are training-data ablations: a two-task model trained on ASR and emotion, and single-task models trained on ASR, emotion, or reasoning alone.
All configurations share the bridge, the adaptation scope, the learning rates, and the batch size.
Appendix~\ref{app:training} lists the training length and learning-rate schedule of each configuration.

\paragraph{Optimization.}
We optimize with AdamW at learning rates of $10^{-4}$ for the bridge and $2\times10^{-5}$ for decoder attention, a linear warm-up over 5,000 updates, and an effective batch size of 256.
Appendix~\ref{app:training} provides the remaining settings.

\subsection{Evaluation Benchmarks and Metrics}
\label{sec:evaluation}

Decoding is greedy throughout, and Appendix~\ref{app:eval} gives the prompts, output limits, scoring rules, and the construction of spoken BBH.

\paragraph{ASR.}
We report WER on the LibriSpeech test-clean and test-other splits, which contain 2,620 and 2,939 utterances.

\paragraph{Emotion Recognition.}
We report the accuracy of the generated completion on our test recordings of ESD, IEMOCAP, and CREMA-D, the three corpora with the most test recordings, and give the other corpora in Appendix~\ref{app:results}.
The analyses in Section~\ref{sec:analysis} use the whole emotion test split.

\paragraph{Reasoning.}
We report answer accuracy on spoken BBH \citep{suzgun2023,chen2026}, SpeechMMLU (SpMMLU) \citep{tan2025}, and MMAU \citep{sakshi2025}.
In these three benchmarks the text prompt contains the written question, and the audio carries the spoken question or the audio clip that the question refers to.

\paragraph{Text-Only Capabilities.}
To assess text-only capabilities after speech adaptation, we compare C-Gate with the base Qwen2.5-7B-Instruct model.
Examples, prompts, shot counts, decoding settings, parsers, and seeds are matched within each benchmark.
We report prompt-level strict IFEval accuracy \citep{zhou2023} and 0-shot normalized HellaSwag accuracy \citep{zellers2019}.

\section{Main Results}
\label{sec:results}

We report the experimental results in terms of word error rate (LibriSpeech test-clean and test-other) and accuracy (emotion recognition and reasoning benchmarks) in Table~\ref{tab:1}.
Appendix~\ref{app:results} provides additional results, including the LibriSpeech WERs of published speech LLMs and emotion accuracy for every corpus and by class.

\begin{table}[t]
\centering
\caption{\textbf{Overall Task Performance.}
WER and accuracy are reported in percent.
The ESD, IEMOCAP, and CREMA-D columns use the test recordings of our train/test split of each corpus.}
\label{tab:1}
{\papertablestyle
\begin{tabular}{@{}*{3}{>{\centering\arraybackslash}p{4.25em}}*{8}{c}@{}}
\toprule
\multicolumn{3}{c}{Training data} & \multicolumn{2}{c}{ASR} & \multicolumn{3}{c}{Emotion} & \multicolumn{3}{c}{Reasoning} \\ \cmidrule(lr){1-3}\cmidrule(lr){4-5}\cmidrule(lr){6-8}\cmidrule(lr){9-11}
ASR & Emotion & Reasoning & Clean & Other & ESD & IEMOCAP & CREMA-D & BBH & SpMMLU & MMAU \\
\midrule
$\checkmark$ & $\times$ & $\times$ & 3.81 & 7.39 & -- & -- & -- & -- & -- & -- \\
$\times$ & $\checkmark$ & $\times$ & -- & -- & 95.4 & 63.4 & 75.0 & -- & -- & -- \\
$\times$ & $\times$ & $\checkmark$ & -- & -- & -- & -- & -- & 28.6 & 53.2 & 44.0 \\
\midrule
$\checkmark$ & $\checkmark$ & $\times$ & 4.36 & 8.34 & 95.7 & 63.8 & 75.5 & -- & -- & -- \\
$\checkmark$ & $\checkmark$ & $\checkmark$ & 3.98 & 8.42 & 94.7 & 64.6 & 74.2 & 43.5 & 61.4 & 48.3 \\
\bottomrule
\end{tabular}
}
\end{table}

\subsection{Overall Task Performance}
\label{sec:overall}

\paragraph{Semantic and Acoustic Tasks.}
We evaluate ASR and emotion recognition as the representative semantic and acoustic tasks.
C-Gate, trained jointly on ASR, emotion recognition, and reasoning (last row of Table~\ref{tab:1}), attains 3.98\% and 8.42\% WER on LibriSpeech test-clean and test-other and emotion accuracy of 94.7\% on ESD, 64.6\% on IEMOCAP, and 74.2\% on CREMA-D.
In both cases, one decoder generates the transcripts and the emotion labels from mixtures of 16 frozen token embeddings per frame.
This shows that continuous mixtures of fixed text embeddings support a semantic and an acoustic task under partial decoder adaptation.

\paragraph{Effect of Joint Training.}
We compare C-Gate with models trained on subsets of the three tasks.
For ASR, the WER of the joint models stays within 0.55 points of the ASR-only model on test-clean and 1.03 points on test-other.
For emotion recognition, the three models trained on emotion data agree within 1.3 points on ESD, IEMOCAP, and CREMA-D, and their accuracy over the whole emotion test split is 84.0\%, 84.1\%, and 84.3\%.
The jointly trained models therefore score close to the single-task models on both tasks, and Section~\ref{sec:analysis} examines how the bridge sequence carries both.

\paragraph{Reasoning Benchmarks.}
C-Gate scores above the reasoning-only model on all three reasoning benchmarks, with 43.5\% against 28.6\% on BBH.
These benchmarks place the written question in the prompt, so the reasoning columns reflect the decoder's answering behavior after joint training (Appendix~\ref{app:results}).

\subsection{Text-Only Performance after Speech Adaptation}
\label{sec:results-text}

To examine how speech adaptation changes the decoder's text-only capabilities, we compare C-Gate with the base Qwen2.5-7B-Instruct model under matched evaluation settings (Table~\ref{tab:4}).
IFEval \citep{zhou2023} prompt-level strict accuracy decreases by 12.01 percentage points after speech adaptation, while HellaSwag \citep{zellers2019} normalized accuracy improves by 2.26 points.
The input embedding table stays fixed, so both changes come from the adapted attention projections.
Appendix~\ref{app:results} extends this comparison to MMLU and GSM8K.

\begin{table}[t]
\centering
\caption{\textbf{Text-Only Performance after Adaptation.}
Scores are percentages under matched evaluation settings, and differences are percentage points computed before rounding.}
\label{tab:4}
{\papertablestyle
\begin{tabular}{@{}l*{4}{>{\centering\arraybackslash}p{0.15\linewidth}}@{}}
\toprule
Benchmark & Items & Base & Adapted & Difference \\
\midrule
IFEval & 541 & 71.35 & 59.33 & $-$12.01 \\
HellaSwag & 10,042 & 64.30 & 66.56 & $+$2.26 \\
\bottomrule
\end{tabular}
}
\end{table}

\section{Representation and Readout Analysis}
\label{sec:analysis}

We ask where the bridge sequence carries task information and how the decoder reads it.
We examine the mixture weights, the selected tokens, and the decoder's attention to the audio frames, and then intervene on the bridge at fixed model weights.
All interventions use the emotion test split and the first 500 test-clean utterances, on which the unmodified C-Gate attains 84.29\% emotion accuracy and 4.01\% WER.

\subsection{Mixture Concentration}
\label{sec:mixture}

\begin{table}[t]
\centering
\caption{\textbf{Routing Statistics across Tasks.}
Statistics use renormalized weights over the selected top-16 support, averaged over the valid frames of the emotion test split, the first 500 test-clean utterances, or 1,000 VoiceBench BBH questions.
$H(\boldsymbol{\alpha}_t)$ is entropy in nats per frame, $H/\log16$ normalizes it by the maximum entropy, and KL is divergence from the uniform distribution $U_{16}$.
High normalized entropy and low KL indicate nearly uniform weights within the selected support.}
\label{tab:3}
{\papertablestyle
\begin{tabular}{@{}*{3}{>{\centering\arraybackslash}p{4.25em}}l*{3}{c}@{}}
\toprule
\multicolumn{3}{c}{Training data} & \multirow{2}{*}{Input task} & \multicolumn{3}{c}{Routing statistics} \\ \cmidrule(lr){1-3}\cmidrule(lr){5-7}
ASR & Emotion & Reasoning & & $H(\boldsymbol{\alpha}_t)$ & $H/\log16$ & KL \\
\midrule
$\times$ & $\checkmark$ & $\times$ & Emotion & 2.744 & 0.990 & 0.0282 \\
\midrule
$\checkmark$ & $\times$ & $\times$ & ASR & 2.660 & 0.960 & 0.1122 \\
\midrule
\multirow{2}{*}{$\checkmark$} & \multirow{2}{*}{$\checkmark$} & \multirow{2}{*}{$\times$} & Emotion & 2.709 & 0.977 & 0.0633 \\
 & &  & ASR & 2.730 & 0.985 & 0.0428 \\
\midrule
\multirow{3}{*}{$\checkmark$} & \multirow{3}{*}{$\checkmark$} & \multirow{3}{*}{$\checkmark$} & Emotion & 2.688 & 0.970 & 0.0844 \\
 & &  & ASR & 2.673 & 0.964 & 0.0997 \\
 & &  & Reasoning & 2.677 & 0.965 & 0.0959 \\
\bottomrule
\end{tabular}
}
\end{table}

We first examine how much task information the mixture weights carry.
The renormalized mixture weights are close to uniform in every training configuration and for every input task (Table~\ref{tab:3}).
Normalized entropy ranges from 0.960 to 0.990, and the bridge vectors of C-Gate have a mean cosine similarity of 0.94 to 0.95 with the unweighted mean of their 16 selected embeddings.
To measure what this residual variation contributes, we set every weight to exactly $1/16$ on the same selected tokens, which lowers emotion accuracy to 81.63\% and raises WER to 5.85\% (Table~\ref{tab:11} in Appendix~\ref{app:interventions}).
Appendix~\ref{app:repr} shows that the softmax before truncation is nearly flat as well and that near-uniform weights still pass gradient to the scorer.
The weights therefore carry a small part of the task information, and the model keeps most of its accuracy without any variation in them.
This leaves two main carriers that Sections~\ref{sec:probes} and~\ref{sec:perturbation} examine in turn: the identity of the selected tokens and their order over time.

\subsection{Emotion Probes across Representations}
\label{sec:probes}

We next test whether the identities of the selected tokens alone carry emotion information.
To this end, we train logistic-regression probes on the emotion test split, with speaker-disjoint folds repeated over ten random partitions (Table~\ref{tab:14} in Appendix~\ref{app:repr}).
A bag-of-tokens probe sees only how often each of the 1,000 most frequently selected tokens occurs in a recording's supports, without weights or frame order.
It reaches 63.3\% accuracy, against 57.6\% for mean-pooled Whisper states and 66.8\% for mean-pooled bridge vectors, and exceeds the Whisper probe in every partition.
For reference, the most frequent class accounts for 23.7\% of the recordings.
This readout is not lexical: a label word appears in only 0.08\% of the supports, and the most frequent tokens are unrelated word pieces that occur in nearly every recording.
This shows that the selected token identities alone support a linear emotion readout at least as accurate as that of the pooled encoder states.

\subsection{Task-Conditional Attention}
\label{sec:attention}

\begin{figure}[t]
\centering
\includegraphics[width=0.84\linewidth]{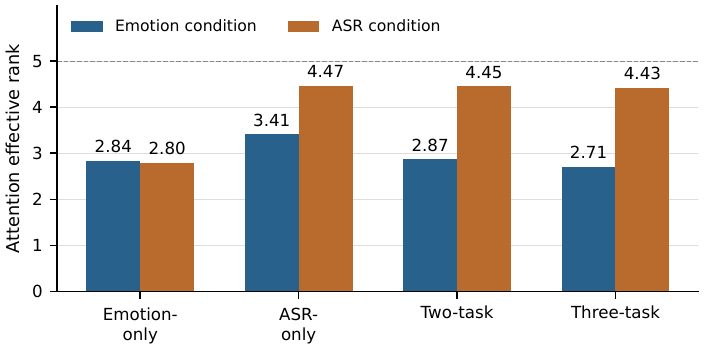}
\caption{\textbf{Decoder Attention to the Audio Frames under Emotion and ASR Inputs.}
The effective rank at decoder layer 14 is higher when the first five answer tokens attend to different frames, and it is at most five (dashed line).}
\label{fig:2}
\end{figure}

We then examine how the decoder attends to the audio frames under each task.
For each example, we feed the evaluation prompt, the bridge frames, and the reference answer, and take the attention from the first five answer tokens to the audio frames at decoder layer 14, averaged over heads.
We summarize this matrix by its entropy-based effective rank \citep{roy2007}, $\exp(-\sum_j p_j\log p_j)$ with $p_j=\sigma_j/\sum_i\sigma_i$ for singular values $\sigma_j$, which is at most five.
A low rank means that the answer tokens share one attention pattern over the audio, and a rank near five means that each token attends to different frames.
When transcribing, the three models trained on ASR reach ranks of 4.43 to 4.47, while the emotion-only model stays at 2.80 on the same inputs (Fig.~\ref{fig:2}).
For emotion completion, the two joint models have ranks of 2.87 and 2.71.
We report layer 14 because layers 14 and 15 are where the transcription rank peaks, and Appendix~\ref{app:repr} summarizes the other layers and the full-transcript ranks.
The joint models thus show position-specific attention when transcribing and more shared attention over the emotion answer.

\subsection{Bridge Mapping and Temporal Sensitivity}
\label{sec:perturbation}

We test whether the decoder's outputs depend on the acoustic input, on the learned correspondence between keys and values, and on frame order, by intervening on the trained C-Gate checkpoint without retraining.

\paragraph{Acoustic Input.}
Replacing the waveform with zeros or with noise lowers emotion accuracy to 17.2\% and 20.0\%, below the 23.7\% share of the most frequent class, and raises WER to 97.7\% and 98.8\% (Appendix~\ref{app:interventions}).
Both tasks therefore depend on the acoustic input.

\begin{figure}[t]
\centering
\includegraphics[width=1.0\linewidth]{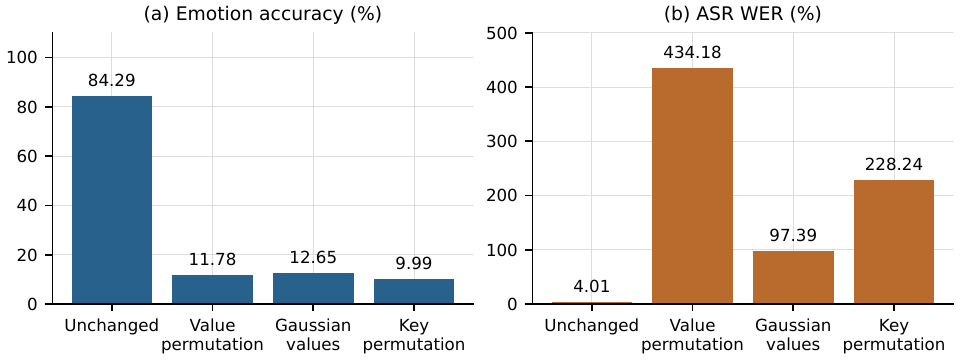}
\caption{\textbf{Performance of C-Gate after Perturbing Its Keys or Values.}
Each perturbation is applied to one trained checkpoint without retraining.}
\label{fig:3}
\end{figure}

\paragraph{Key--Value Mapping.}
We perturb keys or values inside C-Gate while prompt embeddings and the language-model head stay unchanged.
Permuting the values, replacing them with Gaussian vectors, or permuting the keys lowers emotion accuracy to 10--13\% and raises WER above 97\% (Fig.~\ref{fig:3}).
These conditions differ in what they preserve: a row permutation keeps the set of embedding vectors and hence its convex hull, $\operatorname{conv}(\mathbf{P}\mathbf{E})=\operatorname{conv}(\mathbf{E})$ for a permutation matrix $\mathbf{P}$, while Gaussian substitution additionally changes the embedding geometry.
The permutation controls therefore remove only the learned correspondence between a score and the vector it retrieves, and the outputs depend on that correspondence beyond membership in the hull.

\begin{figure}[t]
\centering
\includegraphics[width=1.0\linewidth]{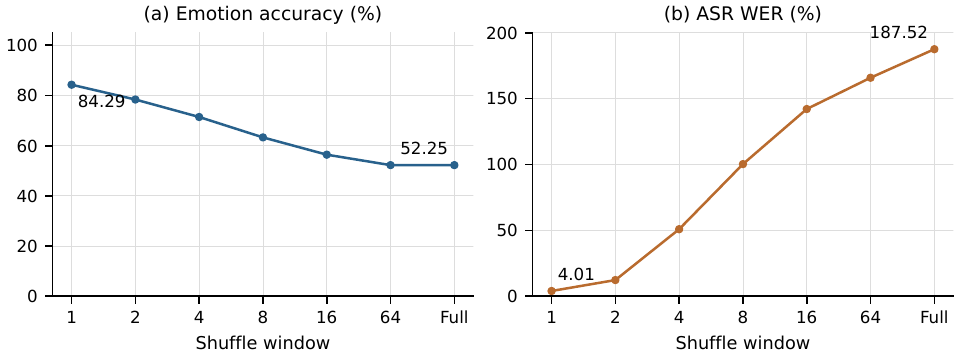}
\caption{\textbf{Performance of C-Gate after Shuffling the Bridge Frames in Time.}
Frames are permuted within non-overlapping windows, where a window of 1 preserves the order and Full shuffles the entire sequence.
Scores are means over three permutation seeds.}
\label{fig:4}
\end{figure}

\paragraph{Temporal Order.}
We permute the bridge frames within non-overlapping windows of increasing size.
Averaged over three permutation seeds, a full shuffle lowers emotion accuracy to 52.25\% and raises WER to 187.52\% (Fig.~\ref{fig:4}).
Transcription fails once local order is lost, with 51\% WER at a window of four frames, while emotion accuracy degrades gradually and stays far above the most frequent class under a full shuffle, in line with the order-free probe of Section~\ref{sec:probes}.

\subsection{Support Size and Pooling Stride}
\label{sec:sensitivity}

\begin{figure}[t]
\centering
\includegraphics[width=\linewidth]{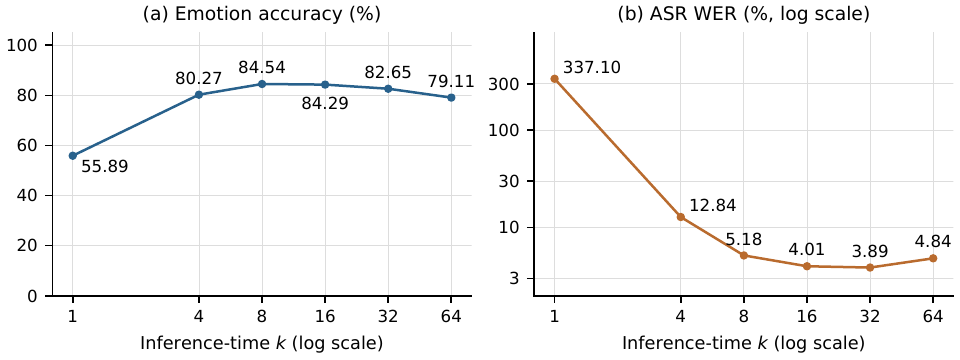}
\caption{\textbf{Changing the Support Size $k$ at Inference.}
Emotion accuracy on the emotion test split and WER on the first 500 test-clean utterances for one C-Gate checkpoint trained with $k=16$.
WER exceeds 100\% at $k=1$ because it counts insertions.}
\label{fig:5}
\end{figure}

Finally, we vary the support size $k$ and the pooling stride at inference in the C-Gate checkpoint trained with $k=16$ and stride 4, without retraining (Fig.~\ref{fig:5}).
Small supports break both tasks: $k=1$ raises WER to 337.10\% and lowers emotion accuracy to 55.89\%.
Around the trained setting, by contrast, the changes are small.
At $k=8$ emotion accuracy is unchanged at 84.54\% and WER rises to 5.18\%, and at $k=32$ WER is unchanged at 3.89\% and emotion accuracy falls to 82.65\%, where unchanged means that the paired 95\% interval includes zero (Appendix~\ref{app:interventions}).
At $k=64$ emotion accuracy falls to 79.11\% and WER rises to 4.84\%.
The stride behaves similarly: changing it from 4 to 2 or 8 degrades both scores, to 14.07\% and 36.50\% WER and to 81.63\% and 70.29\% emotion accuracy (Appendix~\ref{app:interventions}).
For this checkpoint, both tasks therefore depend on mixing several embeddings per frame and on the trained stride.

\section{Discussion and Conclusion}
\label{sec:discussion}

We propose C-Gate, a speech--LLM bridge that represents each speech frame as a sparse convex mixture of frozen text embeddings.
The mixture construction fixes the available embedding values, while task supervision learns which values to select and combine.
With a partially adapted decoder, C-Gate supports both transcription and emotion recognition from one frozen embedding table.
Our analysis shows that the mixture weights stay near uniform, and replacing them with exactly uniform weights costs 2.7 points of emotion accuracy and 1.8 points of WER, so the speech-dependent variation lies mainly in which tokens are selected and in their order.
In particular, token identities alone support a linear emotion readout at least as accurate as that of the pooled encoder states, and the decoder's outputs degrade when frame order or the key--value correspondence is disturbed.

\paragraph{Limitations and Future Work.}
Our study uses one encoder--LLM pair and one training run per configuration.
Matched ablations of the convex-hull constraint and of attention adaptation are left to future work, and Appendix~\ref{app:components} summarizes the evidence behind each design choice.
On the evaluation side, speech adaptation lowers IFEval and MMLU accuracy, and the reasoning scores depend little on the audio in the tested diagnostic.
Finally, scoring the full vocabulary takes about $2.9\times10^{10}$ multiply--accumulate operations per 30-second input (Appendix~\ref{app:cost}), and shortlist scoring is left to future work.

\paragraph{Implications and Broader Impact.}
Fixed embedding mixtures provide an inspectable interface for studying how language models use acoustic information.
Joint speech understanding could support accessible voice interfaces, while emotion inference from acted speech can be unreliable in natural settings and can enable intrusive speaker profiling.
More broadly, speech quality, text retention, and audio contribution are best evaluated together.

\FloatBarrier
\begin{ack}
The work of Zhizheng Wu is supported by the National Natural Science Foundation of China under Grant 62376237.
This work is supported by the Program for Guangdong Introducing Innovative and Entrepreneurial Teams, Grant No.~2023ZT10X044.
The authors have no financial relationships relevant to this work with any entity other than their listed affiliations.
\end{ack}

\clearpage
\appendix
\section{Training Details}
\label{app:training}

This appendix describes what was trained for each row of Table~\ref{tab:1}, including the adaptation scope, the training data, the training configurations, and the optimization settings.

\begin{table}[!htbp]
\centering
\caption{\textbf{Trainable and Frozen Parameters.}
Counts cover the Whisper-large-v3 encoder, the C-Gate scorer, and Qwen2.5-7B-Instruct.}
\label{tab:params}
{\papertablestyle
\begin{tabular}{@{}lrl@{}}
\toprule
Component & Parameters & Status \\
\midrule
C-Gate scorer & 2,492,417 & Trained \\
Decoder attention projections, layers 0--23 & 704,753,664 & Trained \\
\midrule
Decoder MLPs and normalization, layers 0--23 & 4,888,633,344 & Frozen \\
Decoder layers 24--27 and final normalization & 932,234,752 & Frozen \\
Input embedding table & 544,997,376 & Frozen \\
Language-model head & 544,997,376 & Frozen \\
Whisper-large-v3 encoder & 636,968,960 & Frozen \\
\midrule
Total & 8,255,077,889 & 707,246,081 trained \\
\bottomrule
\end{tabular}
}
\end{table}

\subsection{Adaptation Scope}
Table~\ref{tab:params} lists the trainable and frozen parameters.

\subsection{Training Data}

\paragraph{ASR.}
ASR uses the 281,241 utterances of LibriSpeech train-clean-100, train-clean-360, and train-other-500.

\paragraph{Emotion Recognition.}
The emotion pool draws on the English part of CAMEO \citep{christop2025}, which comprises CREMA-D \citep{cao2014}, EMNS \citep{noriy2023}, eNTERFACE \citep{martin2006}, JL-Corpus \citep{james2018}, and RAVDESS \citep{livingstone2018}, together with ESD English \citep{zhou2021}, IEMOCAP \citep{busso2008}, SAVEE \citep{jackson2014}, TESS \citep{dupuis2011}, and MELD audio \citep{poria2019}.
We map the source labels to eight classes and split the 55,398 recordings of the pool by recording into a training portion of 53,251 recordings and 47.55 hours and a test portion of 2,147 recordings and 1.95 hours.
After removing recordings longer than 30 seconds, 53,246 recordings are used for training.
MELD and RAVDESS are used for training only, so the emotion test split consists of the 2,063 test recordings of CREMA-D, EMNS, eNTERFACE, JL-Corpus, ESD, IEMOCAP, SAVEE, and TESS, spoken by 132 speakers.

\paragraph{Reasoning.}
The reasoning sources are ARC \citep{clark2018}, CommonsenseQA \citep{talmor2019}, GSM8K \citep{cobbe2021}, MATH \citep{hendrycks2021b}, MathInstruct \citep{yue2023}, MMLU-Pro \citep{wang2024b}, NuminaMath \citep{li2024}, OpenMath2 \citep{toshniwal2024}, and OrcaMath \citep{mitra2024}, with every question synthesized as speech.
They contain 1,312,333 examples, of which 1,286,024 remain after removing empty answers and audio longer than 30 seconds.
The reasoning audio lasts at least 712 hours.

\begin{table}[!htbp]
\centering
\caption{\textbf{Training Configurations behind Table~\ref{tab:1}.}
Steps count training-loop iterations, and four iterations make one optimizer update.
Each configuration is trained once.}
\label{tab:configs}
{\papertablestyle
\begin{tabular}{@{}llr@{}}
\toprule
Configuration & Training tasks & Evaluated step \\
\midrule
ASR-only & ASR & 100,000 \\
Emotion-only & Emotion & 85,000 \\
Reasoning-only & Reasoning & 100,000 \\
Two-task & ASR, emotion & 90,000 \\
Three-task & ASR, emotion, reasoning & 60,000 \\
\bottomrule
\end{tabular}
}
\end{table}

\subsection{Training Configurations}
Table~\ref{tab:configs} gives the evaluated checkpoint of each configuration.
All configurations share the bridge, the adaptation scope, the learning rates, and the batch size.
They differ in the training tasks, in training length, and in the maximum target length, which is 384 tokens for the three-task model and 256 tokens otherwise.
The reasoning-only model was trained on an earlier reasoning pool that additionally contains OpenHermes 2.5 instruction data \citep{teknium2023} rendered as speech.

\subsection{Optimization}
We use AdamW with coefficients $(0.9, 0.999)$, weight decay 0.01, and gradient clipping at 1.0.
The learning rate is $10^{-4}$ for the bridge and $2\times10^{-5}$ for decoder attention, with a linear warm-up over 5,000 optimizer updates followed by a cosine schedule whose horizon is 300,000 optimizer updates for the joint configurations and 100,000 for the single-task configurations.
Each training-loop iteration processes four micro-batches of two examples on each of eight GPUs, and the optimizer is updated every four iterations, which gives the effective batch size of 256.
Training uses bfloat16 mixed precision.
The data sampler is seeded.

\subsection{Task-Loss Reweighting}
Multi-task configurations combine the per-task cross-entropy losses as
\begin{equation}
\mathcal{L}^{(s)}=\sum_{i\in\mathcal T}w_i^{(s)}\mathcal{L}_i^{(s)},\qquad
w_i^{(s)}\propto\left(\frac{\mathcal{L}_i^{\mathrm{EMA},s}}{\mathcal{L}_i^{\mathrm{init}}}\right)^\alpha,
\label{eq:8}
\end{equation}
where $s$ indexes training-loop iterations and $\mathcal T$ is the task set.
A task whose exponential moving average (EMA) loss has fallen less, relative to its initial loss, receives a larger weight.
The rule is inspired by dynamic task weighting \citep{liu2019} and replaces that method's adjacent-epoch ratio and softmax normalization with an EMA-to-initial-loss ratio.
We use $\alpha=1$, EMA decay 0.9, a 1,000-step warm-up, and weights clipped to $[0.2, 5.0]$.
Single-task configurations optimize their cross-entropy loss directly.

\subsection{Compute}
Training C-Gate to the reported checkpoint took 18.75 hours on eight H100 80\,GB GPUs, about 150 GPU-hours.
The reasoning-only model was trained on eight A100 40\,GB GPUs.
Evaluating one model on both LibriSpeech splits takes 34 to 44 minutes on one A100 80\,GB GPU.

\section{Evaluation Protocols}
\label{app:eval}

This appendix describes how each reported number is produced.
Table~\ref{tab:5} lists the evaluation sets.

\begin{table}[!htbp]
\centering
\caption{\textbf{Evaluation Sets.}
Every row uses the checkpoints in Table~\ref{tab:configs} unless a step is given.}
\label{tab:5}
{\papertablestyle
\begin{tabularx}{0.95\linewidth}{@{}p{0.36\linewidth}X@{}}
\toprule
Evaluation & Items \\
\midrule
ASR & LibriSpeech test-clean, 2,620 utterances, and test-other, 2,939 utterances. \\
Emotion test split & 2,063 test recordings of eight corpora, with ESD, IEMOCAP, and CREMA-D in Table~\ref{tab:1}. \\
Spoken BBH & 4,392 questions in 22 categories. \\
SpeechMMLU and MMAU & 8,549 and 1,000 questions. \\
Text-only benchmarks & The same items for the base and adapted models. \\
Routing statistics & The emotion test split, the first 500 test-clean utterances, and 1,000 VoiceBench BBH questions. \\
Emotion probes & The emotion test split. \\
Fixed-checkpoint interventions & The emotion test split and the first 500 test-clean utterances. \\
Reasoning audio replacement & 1,000 VoiceBench BBH questions, C-Gate at 70,000 steps. \\
\bottomrule
\end{tabularx}
}
\end{table}

\subsection{ASR}
The instruction is ``Transcribe the following speech.''
Decoding is greedy with at most 256 new tokens.
The prompt embeddings precede the audio embeddings.
Audio is sampled at 16\,kHz and truncated to 30 seconds, which affects nine test-clean and seven test-other utterances whose full reference transcripts stay in the denominator.
We uppercase references and predictions, split on whitespace, and divide the total number of word errors by the total number of reference words.
For C-Gate, applying Whisper's English text normalizer instead gives 3.83\% and 8.23\% WER, and excluding the truncated utterances gives 3.88\% and 8.21\%.
The teacher-forced WER in Table~\ref{tab:8} appends the reference tokens to the same prompt and audio, predicts each token from its reference prefix, and scores the decoded argmax sequence.

\subsection{Emotion Recognition}
The instruction is ``The speaker sounds'', which matches the training target ``The speaker sounds \{emotion\}.''
Decoding is greedy with at most 32 new tokens.
We lowercase the output and take the first whole-word occurrence of a label or alias, where the labels are angry, happy, sad, neutral, fearful, disgusted, surprised, and calm, and the aliases are anger, happiness, sadness, fear, disgust, and surprise.
Outputs without a label count as incorrect.

\subsection{Reasoning}
For spoken BBH, SpeechMMLU, and MMAU, the text prompt contains the written question, with answer options where the task has them, followed by an answer-format instruction.
The audio carries the spoken question for BBH and SpeechMMLU and the audio clip under question for MMAU.
Generation is limited to 8 tokens for yes/no, true/false, and valid/invalid BBH categories, 32 tokens for free-form BBH categories, and 64 tokens for multiple-choice questions.
Answer extraction is category-specific for BBH and letter-based for multiple-choice questions.
Our SpeechMMLU protocol scores generated answers, whereas the benchmark's source scores choice likelihoods \citep{tan2025}.

\subsection{Spoken BBH}
Spoken BBH combines the 1,000 questions of the four VoiceBench BBH categories, which are hyperbaton, navigate, sports understanding, and web of lies, with 3,392 questions from 18 further BBH categories \citep{suzgun2023,chen2026}.
For the further categories, we take the question text and reference answers from the BBH release, filter for questions that can be read aloud, and synthesize the audio with Edge TTS using the \texttt{en-US-AriaNeural} voice.
Accuracy and parse rate weight the 4,392 questions equally.

\subsection{Text-Only Benchmarks}
The base and adapted models are evaluated through the text path of the decoder with lm-evaluation-harness 0.4.5 \citep{eleutherai2024}.
Examples, chat template, shot counts, decoding, parsers, and seeds are identical for the two models.

\section{Additional Results}
\label{app:results}

This appendix expands the results of Section~\ref{sec:results} with a comparison to published speech LLMs, per-corpus and per-class emotion results, the reasoning controls, further text-only benchmarks, and a comparison of adaptation scopes.

\subsection{ASR with Limited Adaptation Data}
\begin{table}[!htbp]
\centering
\caption{\textbf{LibriSpeech WER (\%) of C-Gate and Published Speech LLMs.}
Reference scores are copied from the original reports, which differ in training data, backbone, and text normalization.}
\label{tab:2}
{\papertablestyle
\begin{tabular}{@{}lrr@{}}
\toprule
System & test-clean & test-other \\
\midrule
Qwen-Audio & 2.0 & 4.2 \\
Qwen2-Audio & 1.6 & 3.6 \\
Qwen2.5-Omni-7B & 1.8 & 3.4 \\
Kimi-Audio & 1.28 & 2.42 \\
Audio Flamingo 3 & 1.57 & 3.13 \\
WavLLM & 2.0 & 4.8 \\
SALMONN & 2.1 & 4.9 \\
AlignFormer & 3.52 & 6.47 \\
\midrule
C-Gate & 3.98 & 8.42 \\
\bottomrule
\end{tabular}
}
\end{table}

Table~\ref{tab:2} places C-Gate beside published speech LLMs to give scale context for its WER.
C-Gate starts from pretrained Whisper-large-v3 and Qwen2.5-7B-Instruct and adapts them with 960 hours of transcribed speech, together with the emotion and spoken-reasoning data, by updating the bridge and the decoder attention projections.
Two of the reference systems add a large-scale audio pretraining stage: Kimi-Audio pretrains on approximately 13 million hours of raw audio \citep{kimiteam2025}, and the second pretraining stage of Qwen2.5-Omni-7B includes 300 billion audio-related tokens \citep{xu2025}.
A fixed-codebook interface therefore supports transcription without an audio pretraining stage beyond the pretrained encoder, within 0.46 points of the nearest reference system on test-clean and 1.95 points on test-other.

The reference rows are copied from the original reports.
The Qwen-Audio, Qwen2-Audio, Qwen2.5-Omni-7B, Kimi-Audio, and Audio Flamingo 3 entries come from their technical reports \citep{chu2023,chu2024,xu2025,kimiteam2025,goel2025}.
The WavLLM, SALMONN, and AlignFormer entries come from their papers \citep{hu2024,tang2024,fan2025}.
Qwen2-Audio uses its original-framework results, SALMONN its 13B model, and AlignFormer its audio-first configuration before supervised fine-tuning.
These systems differ from C-Gate in training data, backbone, adaptation scope, and text normalization.

\subsection{Emotion Accuracy by Corpus and Class}
\begin{table}[!htbp]
\centering
\caption{\textbf{Emotion Accuracy by Corpus on the Emotion Test Split.}
Accuracy (\%) of the three C-Gate models trained with emotion data on the test recordings of our train/test split of each corpus.}
\label{tab:emotion-corpus}
{\papertablestyle
\begin{tabular}{@{}lrrrr@{}}
\toprule
Corpus & Recordings & Emotion-only & Two-task & Three-task \\
\midrule
ESD & 892 & 95.4 & 95.7 & 94.7 \\
IEMOCAP & 478 & 63.4 & 63.8 & 64.6 \\
CREMA-D & 388 & 75.0 & 75.5 & 74.2 \\
TESS & 132 & 100.0 & 98.5 & 100.0 \\
eNTERFACE & 56 & 94.6 & 98.2 & 96.4 \\
EMNS & 53 & 86.8 & 84.9 & 100.0 \\
JL-Corpus & 49 & 95.9 & 85.7 & 93.9 \\
SAVEE & 15 & 60.0 & 73.3 & 80.0 \\
\midrule
All & 2,063 & 84.0 & 84.1 & 84.3 \\
\bottomrule
\end{tabular}
}
\end{table}

\begin{table}[!htbp]
\centering
\caption{\textbf{Emotion Recall by Class on the Emotion Test Split.}
Recall (\%) within each reference class for the three C-Gate models trained with emotion data.
The calm class does not occur in the test split.}
\label{tab:15}
{\papertablestyle
\begin{tabular}{@{}lrrrr@{}}
\toprule
Class & Recordings & Emotion-only & Two-task & Three-task \\
\midrule
Angry & 488 & 86.7 & 85.5 & 84.0 \\
Disgusted & 90 & 78.9 & 66.7 & 68.9 \\
Fearful & 96 & 63.5 & 72.9 & 68.8 \\
Happy & 432 & 81.7 & 85.4 & 83.8 \\
Neutral & 387 & 82.7 & 83.7 & 82.4 \\
Sad & 354 & 88.1 & 83.9 & 89.8 \\
Surprised & 216 & 88.9 & 91.7 & 93.5 \\
\midrule
Accuracy & 2,063 & 84.0 & 84.1 & 84.3 \\
\bottomrule
\end{tabular}
}
\end{table}

Table~\ref{tab:emotion-corpus} reports accuracy on the emotion test split for every corpus, including the three shown in Table~\ref{tab:1}.
On every corpus with more than 100 recordings the three models agree within two points, and accuracy on these corpora ranges from 63--65\% on IEMOCAP to at least 94.7\% on ESD and TESS.
Over the whole test split, accuracy is 84.0\%, 84.1\%, and 84.3\% for the emotion-only, two-task, and three-task models.
Table~\ref{tab:15} reports recall by class.
Recall is lowest for the disgusted and fearful classes in all three models, and the three-task model has the highest recall for sad and surprised.

\subsection{Reasoning and Answer Extraction}
\begin{table}[!htbp]
\centering
\caption{\textbf{Spoken BBH Accuracy and Answer Extraction.}
Scores are percentages over 4,392 questions.
Parsed accuracy counts only outputs from which an answer could be extracted.
Majority is the accuracy of always giving the most frequent answer of each category.}
\label{tab:6}
{\papertablestyle
\begin{tabular}{@{}lrrr@{}}
\toprule
Model & Accuracy & Parse rate & Parsed accuracy \\
\midrule
Reasoning-only & 28.6 & 79.1 & 36.1 \\
Three-task & 43.5 & 99.2 & 43.9 \\
\midrule
Majority & 33.4 & -- & -- \\
\bottomrule
\end{tabular}
}
\end{table}

\begin{table}[!htbp]
\centering
\caption{\textbf{Replacing the Audio on VoiceBench BBH.}
Accuracy (\%) of C-Gate at 70,000 steps on 1,000 questions, 250 per category, with the written question kept in the text prompt.
All four categories have two answer options.
Majority is the accuracy of always giving the most frequent answer of each category.}
\label{tab:13}
{\papertablestyle
\begin{tabular}{@{}lrrrrrr@{}}
\toprule
Audio input & Overall & Parse rate & Hyperbaton & Navigate & Sports & Web of lies \\
\midrule
Spoken question & 57.9 & 99.3 & 60.0 & 57.2 & 63.6 & 50.8 \\
Noise matched to the waveform level & 54.8 & 97.7 & 50.0 & 56.8 & 63.6 & 48.8 \\
Noise at a fixed level & 55.0 & 97.9 & 50.0 & 56.4 & 63.2 & 50.4 \\
Zeros & 55.4 & 99.5 & 49.2 & 57.6 & 64.0 & 50.8 \\
\midrule
Majority & 53.7 & -- & 51.6 & 58.0 & 54.0 & 51.2 \\
\bottomrule
\end{tabular}
}
\end{table}

Table~\ref{tab:6} separates answer extraction from correctness on spoken BBH.
The reasoning-only model often starts an explanation that the generation limit cuts off, so 20.9\% of its outputs have no extractable answer, against 0.8\% for the three-task model.
Among parsed outputs the two models score 36.1\% and 43.9\%, and always giving the most frequent answer of each category scores 33.4\%.

Table~\ref{tab:13} tests how much the answers depend on the audio when the written question remains in the prompt.
We replace the waveform with Gaussian noise scaled to the level of the original waveform, Gaussian noise at a fixed level, or zeros.
This diagnostic uses a different instruction format from the benchmark runs of Table~\ref{tab:1}, so its absolute accuracy is not comparable with that table.
Spoken questions score 57.9\%, replaced audio scores 54.8--55.4\%, and the majority answer scores 53.7\%.
The difference is confined to hyperbaton, an adjective-ordering task, where accuracy falls from 60.0\% to about 50\%, and the three yes/no categories change by at most 2.0 points.
These observations locate the reasoning scores in answer formatting and in the written question.

\subsection{Text-Only Benchmarks}
\begin{table}[!htbp]
\centering
\caption{\textbf{Additional Text-Only Results for Base and Speech-Adapted Qwen2.5-7B-Instruct.}
Scores are percentages, and differences are percentage points computed before rounding.
Items are prompts for IFEval.
The reasoning training data contain GSM8K and MMLU-Pro.}
\label{tab:18}
{\papertablestyle
\begin{tabular}{@{}lrrrr@{}}
\toprule
Benchmark and metric & Items & Base & Adapted & Difference \\
\midrule
MMLU, 5-shot accuracy & 14,042 & 73.35 & 69.14 & $-$4.22 \\
GSM8K, 5-shot, flexible answer extraction & 1,319 & 69.45 & 69.52 & $+$0.08 \\
GSM8K, 5-shot, strict answer format & 1,319 & 8.11 & 68.69 & $+$60.58 \\
IFEval, instruction-level strict accuracy & 541 & 79.62 & 68.71 & $-$10.91 \\
\bottomrule
\end{tabular}
}
\end{table}

Table~\ref{tab:18} extends Table~\ref{tab:4}.
MMLU \citep{hendrycks2021a} accuracy falls by 4.22 points after adaptation, and GSM8K \citep{cobbe2021} accuracy with flexible answer extraction is unchanged.
Both benchmarks are related to the reasoning training data, which contain GSM8K and MMLU-Pro, so they do not measure retention on unseen material.
The strict GSM8K metric requires the ``\#\#\#\# answer'' format of the GSM8K references, which the base model rarely produces and the adapted model has been trained on.
The instruction-level IFEval metric falls by 10.91 points, in line with the prompt-level metric in Table~\ref{tab:4}.

\subsection{Adaptation Scope under an Earlier Recipe}
\begin{table}[!htbp]
\centering
\caption{\textbf{Adaptation Scope under an Earlier Training Recipe.}
Both models use the C-Gate bridge and the earlier recipe described in this appendix, and differ only in which decoder parameters of layers 0--23 are trained.
Scores are percentages, with AR and TF denoting autoregressive and teacher-forced WER on LibriSpeech test-clean and test-other.
Emotion is accuracy on the emotion test split.}
\label{tab:8}
{\papertablestyle
\begin{tabular}{@{}l*{8}{c}@{}}
\toprule
\multirow{2}{*}{Trained decoder parameters} & \multicolumn{2}{c}{AR WER} & \multicolumn{2}{c}{TF WER} & \multirow{2}{*}{Emotion} & \multirow{2}{*}{BBH} & \multirow{2}{*}{SpMMLU} & \multirow{2}{*}{MMAU} \\ \cmidrule(lr){2-3}\cmidrule(lr){4-5}
 & Clean & Other & Clean & Other & & & & \\
\midrule
All parameters & 4.51 & 9.35 & 4.08 & 7.99 & 66.8 & 37.4 & 68.2 & 47.3 \\
Attention projections & 5.70 & 10.62 & 5.40 & 9.48 & 53.8 & 40.1 & 72.3 & 50.3 \\
\bottomrule
\end{tabular}
}
\end{table}

Table~\ref{tab:8} compares two models that use the C-Gate bridge and differ only in the trained decoder parameters of layers 0--23.
One model trains all parameters of these layers, 5.59 billion in total, and the other trains the attention projections, as C-Gate does.
Both come from an earlier recipe that differs from C-Gate's in data and schedule.
It uses 460 hours of LibriSpeech, the emotion training recordings, spoken Alpaca instructions, and text-only replay with fixed sampling ratios, includes no reasoning speech and no loss reweighting, and is evaluated at 300,000 steps.
Under this recipe, training all decoder parameters gives lower WER and higher emotion accuracy, and training attention projections gives higher scores on the three reasoning benchmarks.

\section{Routing and Representation Diagnostics}
\label{app:repr}

This appendix gives the protocols and full results behind Sections~\ref{sec:mixture}--\ref{sec:attention}.

\subsection{Routing Statistics}
\begin{table}[!htbp]
\centering
\caption{\textbf{Support Statistics beyond the Renormalized Weights.}
Means over valid frames.
Full entropy is the entropy of the softmax over all 152,064 embedding rows divided by its maximum.
Mass is the probability that this softmax assigns to the 16 selected entries, in units of $10^{-4}$, where a uniform distribution assigns 1.05.
Cosine compares each bridge vector with the unweighted mean of its selected embeddings.
Overlap is the Jaccard index between the supports of adjacent frames.}
\label{tab:7}
{\papertablestyle
\begin{tabular}{@{}llrrrrr@{}}
\toprule
Model & Input task & Frames & Full entropy & Mass & Cosine & Overlap \\
\midrule
Emotion-only & Emotion & 84,663 & 0.9983 & 2.94 & 0.984 & 0.13 \\
\midrule
ASR-only & ASR & 49,574 & 0.9916 & 9.96 & 0.945 & 0.16 \\
\midrule
Two-task & Emotion & 84,663 & 0.9965 & 4.13 & 0.964 & 0.14 \\
 & ASR & 49,574 & 0.9960 & 4.26 & 0.974 & 0.13 \\
\midrule
Three-task & Emotion & 84,663 & 0.9995 & 2.39 & 0.950 & 0.25 \\
 & ASR & 49,574 & 0.9995 & 2.57 & 0.941 & 0.26 \\
 & Reasoning & 279,819 & 0.9995 & 2.55 & 0.942 & 0.29 \\
\bottomrule
\end{tabular}
}
\end{table}

Table~\ref{tab:3} reports the renormalized weights over the selected support, and Table~\ref{tab:7} adds statistics of the softmax before truncation and of the supports.
Valid frames are the bridge frames that cover the recording, as opposed to the padding of the 30-second encoder window.
All statistics are means over the valid frames of the emotion test split for emotion input, the first 500 test-clean utterances for ASR input, and 1,000 VoiceBench BBH questions for reasoning input.
The softmax over all 152,064 embedding rows is nearly flat before truncation.
Its entropy is at least 99.1\% of the maximum in every row of the table, and the 16 selected entries hold between $2.4\times10^{-4}$ and $1.0\times10^{-3}$ of the probability mass, 2 to 10 times what a uniform distribution assigns to 16 entries.
Truncation and renormalization, more than the sharpness of the softmax, therefore determine the mixture.
Adjacent frames share few selected tokens, with a mean Jaccard overlap between 0.13 and 0.29.

\subsection{Gradient of the Renormalized Weights}
Near-uniform weights could indicate a saturated scorer that no longer receives gradient.
For a fixed support $S_t$, the renormalized weights are a softmax over the selected scores:
\begin{equation}
\alpha_{t,v}=\frac{\exp z_{t,v}}{\sum_{u\in S_t}\exp z_{t,u}},\qquad
\frac{\partial\alpha_{t,v}}{\partial z_{t,u}}=\alpha_{t,v}(\mathbf{1}_{v=u}-\alpha_{t,u}),\quad u,v\in S_t.
\label{eq:7}
\end{equation}
At uniform weights over $k$ entries, this Jacobian has $k-1$ eigenvalues equal to $1/k$, whereas it vanishes when a single weight approaches one.
Near-uniform weights therefore leave the scorer trainable wherever the support is unchanged.

\subsection{Selected Tokens}
On the 84,663 valid frames of the emotion test split, C-Gate selects 19,997 distinct tokens, and 43 of them account for half of all selections.
The most frequent tokens are word pieces unrelated to the input, such as ``CSS'', ``foundation'', ``repeat'', and ``.day'', together with Chinese word pieces, and each of them occurs in nearly every recording.
A token that spells an emotion label or one of its aliases appears in the support of 70 frames, 0.08\% of the total.
The supports thus act as a learned code for the decoder and do not name the emotion.

\subsection{Emotion Probes}
\begin{table}[!htbp]
\centering
\caption{\textbf{Linear Emotion Probes on C-Gate Representations.}
Accuracy (\%) on the emotion test split is the mean $\pm$ standard deviation over ten random speaker-disjoint partitions.}
\label{tab:14}
{\papertablestyle
\begin{tabular}{@{}lrc@{}}
\toprule
Representation & Dimension & Accuracy \\
\midrule
Mean-pooled Whisper states & 1,280 & 57.6 $\pm$ 5.7 \\
Bag of selected token identities & 1,000 & 63.3 $\pm$ 5.2 \\
Mean-pooled bridge output & 3,584 & 66.8 $\pm$ 3.9 \\
PCA of pooled bridge output & 32 & 67.8 $\pm$ 4.0 \\
\midrule
Most frequent class & -- & 23.7 \\
\bottomrule
\end{tabular}
}
\end{table}

Table~\ref{tab:14} reports multinomial logistic-regression probes with regularization parameter $C=1$ on C-Gate.
We assign the 132 speakers of the emotion test split at random to five folds, score every recording once, and repeat this over ten random partitions.
Features are computed over the valid frames of each recording and standardized with training-fold statistics.
The bag-of-tokens features are the relative frequencies of the 1,000 tokens selected most often in the training folds, and the PCA is fitted on the training folds.
The bag exceeds the pooled Whisper states in all ten partitions, by 3.5 to 8.2 points.

\subsection{Attention Rank}
Figure~\ref{fig:2} uses 100 recordings of the emotion test split and 100 test-clean utterances, taken at regular intervals from the sets of Table~\ref{tab:5}.
The input follows the evaluation layout: the chat-formatted instruction, the valid bridge frames, and the teacher-forced reference answer.
The answer is ``The speaker sounds \{emotion\}.'' for emotion, which has five tokens, and the reference transcript for ASR, where three utterances with fewer than five answer tokens are excluded.
Rows of the attention matrix are the first five answer tokens, columns are the audio frames, and attention is averaged over heads without renormalizing the rows.
For the three models trained on ASR, the ASR rank is highest at layers 14 and 15, between 4.43 and 4.56, and averages 3.05 to 3.12 over the 28 layers.
Over all answer tokens of an utterance, 35.7 on average, the ASR rank at layer 14 is 30.1 to 30.5 for these models and 11.1 for the emotion-only model.
Under the emotion condition, the rank at layer 14 is 2.71 to 2.87 for the models trained on emotion and 3.41 for the ASR-only model.

\section{Fixed-Checkpoint Interventions}
\label{app:interventions}

This appendix gives the protocols and values behind Sections~\ref{sec:perturbation} and~\ref{sec:sensitivity}.
Every intervention changes one component of the trained C-Gate checkpoint at inference and keeps the weights, prompts, decoding, and scoring fixed.
All interventions use the 2,063 recordings of the emotion test split and the first 500 test-clean utterances, and unparsed emotion outputs count as incorrect.

\begin{table}[!htbp]
\centering
\caption{\textbf{Replacing the Waveform.}
Scores (\%) of one C-Gate checkpoint on the emotion test split and the first 500 test-clean utterances.}
\label{tab:10}
{\papertablestyle
\begin{tabular}{@{}lrr@{}}
\toprule
Audio input & Emotion accuracy & WER \\
\midrule
Original waveform & 84.29 & 4.01 \\
Zeros & 17.16 & 97.71 \\
Gaussian noise at the waveform level & 20.02 & 98.84 \\
\bottomrule
\end{tabular}
}
\end{table}

\begin{table}[!htbp]
\centering
\caption{\textbf{Replacing the Mixture Weights and Perturbing the Key--Value Mapping.}
Scores (\%) of one C-Gate checkpoint on the emotion test split and the first 500 test-clean utterances, with prompt embeddings and all model weights unchanged.
Differences are relative to the unchanged bridge in percentage points, with paired 95\% block-bootstrap intervals.
Fig.~\ref{fig:3} plots the unchanged bridge and the three mapping conditions.}
\label{tab:11}
{\papertablestyle
\begin{tabular}{@{}lrlrl@{}}
\toprule
C-Gate bridge & Emotion accuracy & Difference [95\% interval] & WER & Difference [95\% interval] \\
\midrule
Unchanged & 84.29 & -- & 4.01 & -- \\
Uniform weights & 81.63 & $-$2.67 [$-$4.88, $-$1.36] & 5.85 & $+$1.84 [$+$1.16, $+$2.46] \\
Value permutation & 11.78 & $-$72.52 [$-$78.76, $-$63.63] & 434.18 & $+$430.17 [$+$364.39, $+$498.07] \\
Gaussian values & 12.65 & $-$71.64 [$-$78.88, $-$61.29] & 97.39 & $+$93.38 [$+$92.74, $+$93.93] \\
Key permutation & 9.99 & $-$74.31 [$-$82.49, $-$62.42] & 228.24 & $+$224.23 [$+$177.51, $+$271.30] \\
\bottomrule
\end{tabular}
}
\end{table}

\begin{table}[!htbp]
\centering
\caption{\textbf{Graded Temporal Shuffling.}
Scores (\%) are means over three permutation seeds on the emotion test split and the first 500 test-clean utterances.
Window 1 preserves order, and Full shuffles the entire sequence.
This table gives the values plotted in Fig.~\ref{fig:4}.}
\label{tab:12}
{\papertablestyle
\begin{tabular}{@{}lrrrrrrr@{}}
\toprule
Window & 1 & 2 & 4 & 8 & 16 & 64 & Full \\
\midrule
Emotion accuracy & 84.29 & 78.40 & 71.43 & 63.31 & 56.42 & 52.25 & 52.25 \\
ASR WER & 4.01 & 12.30 & 50.91 & 100.32 & 142.11 & 165.87 & 187.52 \\
\bottomrule
\end{tabular}
}
\end{table}

\subsection{Audio Replacement}
Table~\ref{tab:10} replaces the waveform with zeros or with Gaussian noise scaled to the root-mean-square level of the original waveform.
Emotion accuracy falls below the 23.7\% share of the most frequent class, and WER approaches 100\%.

\subsection{Mixture Weights and Key--Value Mapping}
Table~\ref{tab:11} changes one component inside C-Gate.
Uniform weights keep the 16 tokens selected for each frame and set every weight to $1/16$, so each bridge vector becomes the unweighted mean of its selected embeddings.
Value permutation applies one random permutation to the embedding rows that the selected indices retrieve.
Key permutation applies it to the rows from which the keys are projected, so that scores select different entries while the values stay in place.
Gaussian values replace the embedding rows with independent Gaussian vectors rescaled to the mean and standard deviation of each embedding dimension.
The ordinary text path of the decoder uses the original embedding table in all conditions.

\subsection{Temporal Shuffling}
Table~\ref{tab:12} permutes the bridge frames within non-overlapping windows, with a separate permutation for each utterance and seed.
The three seeds vary only the permutations.

\begin{table}[!htbp]
\centering
\caption{\textbf{Support-Size Sensitivity without Retraining.}
Scores (\%) use the emotion test split and the first 500 test-clean utterances, with the model trained at $k=16$.
Differences are relative to $k=16$ in percentage points, with paired 95\% block-bootstrap intervals.
This table gives the values plotted in Fig.~\ref{fig:5}.}
\label{tab:16}
{\papertablestyle
\begin{tabular}{@{}crlrl@{}}
\toprule
$k$ & Emotion accuracy & Difference [95\% interval] & WER & Difference [95\% interval] \\
\midrule
1 & 55.89 & $-$28.41 [$-$31.61, $-$24.34] & 337.10 & $+$333.09 [$+$315.24, $+$354.71] \\
4 & 80.27 & $-$4.02 [$-$6.24, $-$2.36] & 12.84 & $+$8.83 [$+$8.00, $+$9.73] \\
8 & 84.54 & $+$0.24 [$-$0.97, $+$1.23] & 5.18 & $+$1.17 [$+$0.68, $+$1.73] \\
16 & 84.29 & -- & 4.01 & -- \\
32 & 82.65 & $-$1.65 [$-$3.58, $-$0.33] & 3.89 & $-$0.12 [$-$0.45, $+$0.18] \\
64 & 79.11 & $-$5.19 [$-$8.92, $-$3.15] & 4.84 & $+$0.83 [$+$0.25, $+$1.57] \\
\bottomrule
\end{tabular}
}
\end{table}

\begin{table}[!htbp]
\centering
\caption{\textbf{Pooling-Stride Sensitivity without Retraining.}
Scores (\%) use the emotion test split and the first 500 test-clean utterances, with the model trained at stride 4.
Frames gives the mean number of bridge frames per ASR utterance.
Differences are relative to stride 4 in percentage points, with paired 95\% block-bootstrap intervals.}
\label{tab:17}
{\papertablestyle
\begin{tabular}{@{}crrlrl@{}}
\toprule
Stride & Frames & Emotion accuracy & Difference [95\% interval] & WER & Difference [95\% interval] \\
\midrule
2 & 197.8 & 81.63 & $-$2.67 [$-$4.93, $-$1.24] & 14.07 & $+$10.06 [$+$8.53, $+$11.69] \\
4 & 99.1 & 84.29 & -- & 4.01 & -- \\
8 & 49.8 & 70.29 & $-$14.01 [$-$16.91, $-$10.28] & 36.50 & $+$32.49 [$+$29.41, $+$35.41] \\
\bottomrule
\end{tabular}
}
\end{table}

\subsection{Support Size and Stride}
Tables~\ref{tab:16} and~\ref{tab:17} change the support size or the pooling stride at inference for the checkpoint trained with $k=16$ and stride 4.
The intervals are paired percentile bootstrap intervals from 10,000 resamples of the 132 speakers of the emotion test split or the 9 LibriSpeech speakers, without adjustment for multiple comparisons.
A stride of 2 doubles the number of bridge frames and a stride of 8 halves it, so the stride conditions change the sequence length as well as the temporal resolution.

\section{Cost of Full-Vocabulary Scoring}
\label{app:cost}

This appendix counts the operations of the bridge.
C-Gate scores every vocabulary entry for every frame before selecting $k$ entries, so selection does not reduce the scoring cost.
With $V=152{,}064$ embedding rows, $d_{\mathrm{llm}}=3{,}584$, $d_p=512$, and $T'=375$ frames for a 30-second input, projecting the keys costs $Vd_{\mathrm{llm}}d_p\approx2.8\times10^{11}$ multiply--accumulate operations and scoring costs $T'Vd_p\approx2.9\times10^{10}$ per input.
The $T'\times V$ score matrix has $5.7\times10^{7}$ entries, which occupy 109\,MiB in bfloat16 and 218\,MiB in 32-bit floating point.
Key projection runs once per batch during training and can be cached at inference, because the keys depend only on fixed weights.
Mixing the selected values adds $T'kd_{\mathrm{llm}}$ operations.
These counts cover the bridge alone and exclude end-to-end latency.

\section{Evidence for Each Design Choice}
\label{app:components}

\begin{table}[!htbp]
\centering
\caption{\textbf{Evidence for Each Design Choice.}}
\label{tab:19}
{\papertablestyle
\begin{tabularx}{0.95\linewidth}{@{}>{\raggedright\arraybackslash}p{0.24\linewidth}>{\raggedright\arraybackslash}X@{}}
\toprule
Design choice & Evidence \\
\midrule
Convex constraint & Weights are nonnegative and sum to one over frozen embedding rows (Eq.~\ref{eq:4}), so hull membership holds by construction. \\
Mixture weights & Setting every weight to 1/16 on the same selected tokens lowers emotion accuracy by 2.7 points and raises WER by 1.8 points (Section~\ref{sec:mixture}). \\
Learned key--value mapping & Permuting keys or values, or replacing values with Gaussian vectors, degrades both tasks (Section~\ref{sec:perturbation}). \\
Temporal order & Windowed shuffles degrade both tasks, more for larger windows, in three seeds (Section~\ref{sec:perturbation}). \\
Support size and stride & Changing $k$ or the stride at inference changes both scores (Section~\ref{sec:sensitivity}). \\
Partial decoder adaptation & Parameter account (Table~\ref{tab:params}) and text-only evaluation (Tables~\ref{tab:4} and~\ref{tab:18}). \\
Loss reweighting & Used in both multi-task configurations (Appendix~\ref{app:training}), with no separate ablation. \\
\bottomrule
\end{tabularx}
}
\end{table}

Table~\ref{tab:19} summarizes the evidence in this paper for each design choice.
The interventions on the mixture weights, the key--value mapping, the temporal order, the support size, and the stride are applied at inference to one trained checkpoint without retraining.

\end{document}